\documentclass[10pt,a4paper]{article}
\usepackage[latin2]{inputenc}
\usepackage{graphicx}
\usepackage{authblk}
\usepackage{ulem}
\usepackage{amsmath}
\usepackage{fancyhdr}

\begin{document}

\title {Gray Level Image Threshold Using Neutrosophic Shannon Entropy}

\author {Vasile Patrascu,\\
 Research Center in Electrical Engineering, Electronics and Information
Technology, \\ Valahia University,\\Targoviste, Romania,\\E-mail: patrascu.v@gmail.com }

\date{}
\maketitle
% apare headerul si pe prima pagina
\thispagestyle{fancy}

\begin{abstract}
This article presents a new method of segmenting grayscale
images by minimizing Shannon's neutrosophic entropy. For the proposed
segmentation method, the neutrosophic information components, i.e., the
degree of truth, the degree of neutrality and the degree of falsity are
defined taking into account the belonging to the segmented regions and
at the same time to the separation threshold area. The principle of the
method is simple and easy to understand and can lead to multiple
thresholds. The efficacy of the method is illustrated using some test
gray level images. The experimental results show that the proposed
method has good performance for segmentation with optimal gray level
thresholds.
\end{abstract}

\textbf{Keywords}: image segmentation, neutrosophic information, Shannon entropy,
gray level image threshold.

\textbf{AMS Classification}: 68U10, 62H35.

\section{Introduction}

Image segmentation is a process that divides the image into its
component parts. One of the most used methods is the thresholding one.
Recent advances in the neutrosophic representation of information allow
different possibilities for the development of new image segmentation
techniques. Neutrosophic models have the ability to work with data
uncertainty and appear as an alternative to improving the threshold
selection process so we get the right segmentation. In this paper we use
as a thresholding function the Shannon entropy of neutrosophic
information.

Next, the structure of the article is the following: Section 2 gives the
neutrosophication procedure for the gray level images. The proposed
neutrosophication is very specific to the thresholding technique;
Section 3 shows how Shannon entropy can be calculated for the
neutrosophic information; Section 4 shows the thresholding algorithm;
Section 5 shows experimental results; Section 6 shows the conclusions
while the last section is the references one.

\section{The Neutrosophic Information Construction}

Neutrosophic representation of information was proposed by Smarandache
[\ref{r2}], [\ref{r7}], [\ref{r8}] and [\ref{r9}] as an extension of fuzzy
representation proposed by Zadeh [\ref{r12}] and intuitionistic fuzzy
representation proposed by Atanassov [\ref{r1}]. Primary neutrosophic
information is defined by three parameters: degree of truth $T$,
degree of falsity $F$ and degree of neutrality $I$. In the next, we
will show, how can construct the neutrosophic information for a gray
level image, in such a way to be useful for threshold methods.

Consider the gray levels in the range $[0,1]$. We denote with $X$
the multiset [\ref{r11}] of the gray levels existing in the whole image
and take a point $t\in(\min (X),max(X))$. Also, we denote with $X_{1}
$ the multiset of gray levels existing in the image from the interval $
[0,t]$ and with $X_{2}$ the multiset of gray levels existing in the
image from the interval $[t,1]$, that is:

\begin{equation}\label{1}
X_{1}=\lbrace x\in X| x\le t\rbrace
\end{equation}

\begin{equation}\label{2}
X_{2}=\lbrace x\in X| x\ge t\rbrace
\end{equation}

The threshold $t$ represents the neutrality point between the two
multisets $X_{1}$ and $X_{2}$. Then, we calculate the average $v_{1}$ for gray levels lower than $t$ and the average $v_{2}$ for gray levels larger than $t$ with formulas:

\begin{equation}\label{3}
v_{1}(t)=\frac{1}{card(X_{1})}\sum_{x\in X_{1}}{x}
\end{equation}

\begin{equation}\label{4}
v_{2}(t)=\frac{1}{card(X_{2})}\sum_{x\in X_{2}}{x}
\end{equation}

For calculation the dissimilarity between two gray levels $x,y\in[0,1]$
we will use the following metric [\ref{r5}], [\ref{r6}]:

$d:\lbrack 0,1\rbrack \times \lbrack 0,1\rbrack \to [0,1],
$

\begin{equation}\label{5}
d(x,y)=\frac{2|x-y|}{1+\vert x-0.5\vert +|y-0.5|}
\end{equation}

Using the metric $d$, we calculate the dissimilarities between each
gray level $x$ and the averages $v_{1}$, $v_{2}$ and the
threshold $t$. It results $d(x,v_{1})$, $d(x,v_{2})$ and $
d(x,t)$.

\begin{equation}\label{6}
d(x,v_{1})=\frac{2|x-v_{1}(t)|}{1+\vert x-0.5\vert +|v_{1}(t)-0.5|}
\end{equation}

\begin{equation}\label{7}
d(x,v_{2})=\frac{2|x-v_{2}(t)|}{1+\vert x-0.5\vert +|v_{2}(t)-0.5|}
\end{equation}

\begin{equation}\label{8}
d(x,t)=\frac{2|x-t|}{1+\vert x-0.5\vert +|t-0.5|}
\end{equation}

We denote the minimum between the dissimilarities $d_{1}(x,v_{1})$ and
$d_{2}(x,v_{2})$ with:

\begin{equation}\label{9}
d_{v}(x)=min(d(x,v_{1}),d(x,v_{2}))
\end{equation}

With these dissimilarities, we will construct the neutrosophic
information associated with each gray level $x$ and related to the
threshold $t$: the degree of truth $T(x,t)$, the degree of falsity
$F(x,t)$ and the degree of neutrality $I(x,t)$.\\

The degree of truth:

\begin{equation}\label{10}
T(x,t)=\frac{d(x,v_{2})-d(x,v_{1})\cdot d(x,v_{2})}{d(x,v_{1})+d(x,v_{2})-d(x,v_{1})\cdot d(x,v_{2})}
\end{equation}

The degree of falsity:

\begin{equation}\label{11}
F(x,t)=\frac{d(x,v_{1})-d(x,v_{1})\cdot d(x,v_{2})}{d(x,v_{1})+d(x,v_{2})-d(x,v_{1})\cdot d(x,v_{2})}
\end{equation}

The degree of neutrality:

\begin{equation}\label{12}
I(x,t)=\frac{d_{v}(x)-d(x,t)\cdot d_{v}(x)} {d(x,t)+d_{v}(x)-d(x,t)\cdot d_{v}(x)}
\end{equation}

\begin{figure}[hb]
\centering
\includegraphics[width=6.78cm,height=5.08cm]{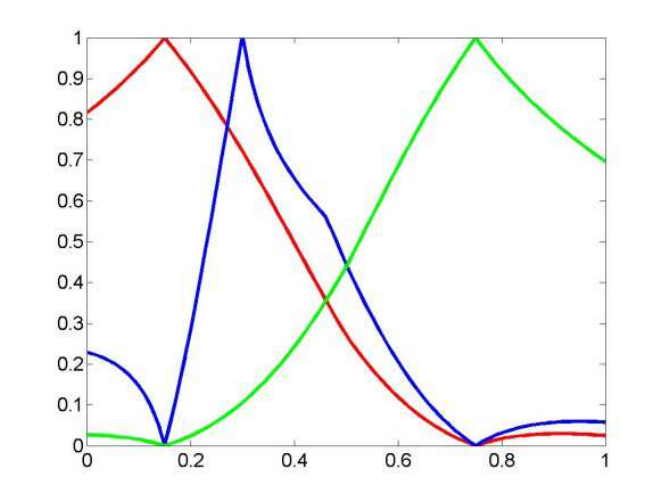}
\caption{The graphic of the functions $T (red),I (blue),F (green)$ for
$v_{1}=0.15$, $v_{2}=0.75$ and $t=0.3$.}
\label{TIF}
\end{figure}

In figure \ref{TIF}, we can see the graphic of the functions $T(x,t)$, $
I(x,t)$ and $F(x,t)$ for the particular case: $v_{1}=0.15$, $
v_{2}=0.75$ and $t=0.3$

\section{The Shannon entropy for neutrosophic information}

In this paper, the Shannon function [\ref{r10}] is used as a measure for
the neutrosophic information uncertainty. We do the following notations:\\

The bifuzzy undefinedness $U$:

\begin{equation}\label{13}
U(x,t)=\max (0,1-T(x,t)-F(x,t))
\end{equation}

The bifuzzy contradiction $C$:

\begin{equation}\label{14}
C(x,t)=\max (0,T(x,t)+F(x,t)-1)
\end{equation}

The escort fuzzy degree of truth $p_{T}$:

\begin{equation}\label{15}
p_{T}(x,t)=\frac{T(x,t)+U(x,t)+\dfrac{I(x,t)}{2}}{1+I(x,t)+U(x,t)+C(x,t)}
\end{equation}

The escort fuzzy degree of falsity $p_{F}$:

\begin{equation}\label{16}
p_{F}(x,t)=\frac{F(x,t)+U(x,t)+\dfrac{I(x,t)}{2}}{1+I(x,t)+U(x,t)+C(x,t)}
\end{equation}

Then, we calculate the neutrosophic Shannon entropy $e(x,t)$ for each
gray level $x$ using formula proposed in [\ref{r3}, \ref{r4}].

\begin{equation}\label{17}
e(x,t)=\frac{p_{T}(x,t)\cdot ln(p_{T}(x,t))+p_{F}(x,t)\cdot ln(p_{F}(x,t))} {-ln(2)}
\end{equation}

In the space $(T,I,F)$ the Shannon entropy for neutrosophic
information verifies the general conditions of neutrosophic uncertainty
[\ref{r4}]:
\\

i) $
e(1,0,0)=e(0,0,1)=0
$\\

ii) $
e(T,I,T)=e(F,I,F)=1
$\\

iii) $
e(T,I,F)=e(F,I,T)
$\\

iv) $
e(T_{1},I_{1},F_{1})\le e(T_{2},I_{2},F_{2})
$ if $ \vert
T_{1}-F_{1}\vert \ge \vert T_{2}-F_{2}\vert$, $ \vert
T_{1}+F_{1}-1\vert \le \vert T_{2}+F_{2}-1\vert $ and $ I_{1}\le I_{2}$.
\\

The property (iv) shows that the Shannon entropy decreases with $\vert
T-F\vert $ , increases with $\vert T+F-1\vert $ and increases with $I
$.

From (iv) it results that $e(T,I,F)\in[0,1]$ because $e(T,I,F)\ge
e(1,0,0)$ and $e(T,I,F)\le e(0,1,0)$. Also, we must mention that
there exists the following equality:

\begin{equation}\label{18}
\vert T+F-1\vert =C+U
\end{equation}

\section{The algorithm for thresholding operation}

We denote by $x_{m}=min(X)$ and $x_{M}=max(X)$. For each $
t\in(x_{m},x_{M})\cap \lbrace \frac{1}{Q},\frac{2}{Q},\ldots
,\frac{Q-1}{Q}\rbrace $, we calculate the entropy average for the three
fuzzy sets defined by the neutrosophic components $T$, $I$ and $F
$. Here, the natural number $Q$ is the step of threshold
quantization. Typically, $Q=255$. It results the following three
partial entropy functions: $e_{T}(t)$, $e_{I}(t)$, $e_{F}(t)$.

\begin{equation}
\label{19}
e_{T}(t)=\frac{\sum_{x\in X}{T(x,t)\cdot e(x,t)}}{\sum_{x\in X}{T(x)}}
\end{equation}

\begin{equation}\label{20}
e_{I}(t)=\frac{\sum_{x\in X}{I(x,t)\cdot e(x,t)}}{\sum_{x\in X}{I(x,t)}}
\end{equation}

\begin{equation}\label{21}
e_{F}(t)=\frac{\sum_{x\in X}{F(x,t)\cdot e(x,t)}}{\sum_{x\in X}{F(x,t)}}
\end{equation}

Then, we define the optimal function that is the total entropy, which is
calculated with the average of the three partial entropies calculated
above.

\begin{equation}\label{22}
E(t)=\frac{e_{T}(t)+e_{I}(t)+e_{F}(t)}{3}
\end{equation}

The segmentation thresholds are the local minimum points of the total
entropy $E$.

\section{Experimental results}

\begin{figure}[hb]
\centering
\begin{tabular}{cc}
{\includegraphics[width=5cm,height=5cm]{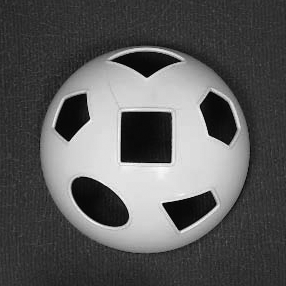}} \ & \
{\includegraphics[width=5cm,height=5cm]{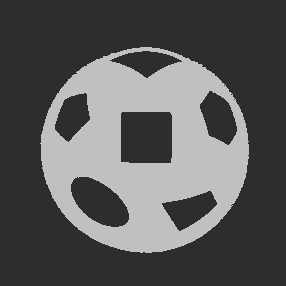}}\\
a) & b)
\end{tabular} 
\caption{The image ball (\textit{a}). The segmented image with two gray levels (\textit{b}).}
\label{ball}
\end{figure}

\begin{figure}[h]
\centering
\includegraphics[width=6.66cm,height=5cm]{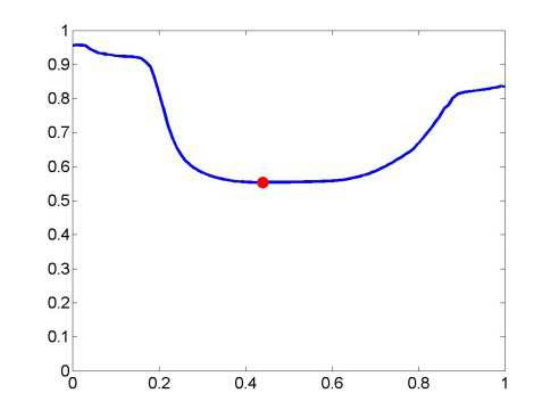}
\caption{The graphic of neutrosophic Shannon entropy where the red
circle represents the gray level threshold (local minimum) for image ball.}
\label{shan_ball}
\end{figure}

\begin{figure}[ht]
\centering
\begin{tabular}{cc}
{\includegraphics[width=5cm,height=5cm]{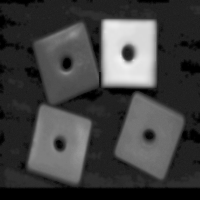}} \ & \
{\includegraphics[width=5cm,height=5cm]{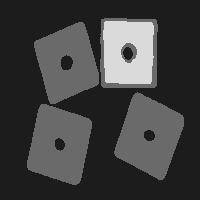}}\\
a) & b)
\end{tabular} 
\caption{The image block (\textit{a}). The segmented image with three gray levels (\textit{b}).}
\label{block}
\end{figure}

\begin{figure}[ht]
\centering
\includegraphics[width=6.66cm,height=5cm]{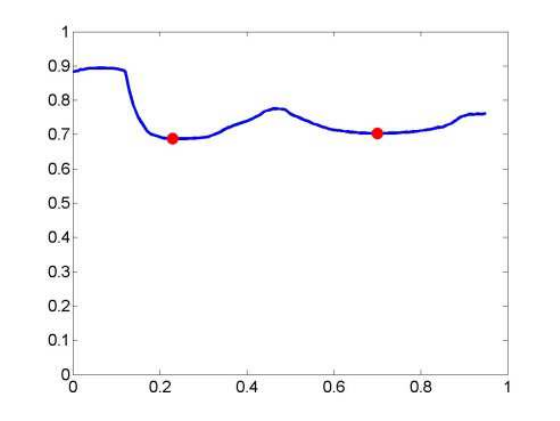}
\caption{The graphic of neutrosophic Shannon entropy where the red
circles represent the two gray level thresholds (local minima) for image block.}
\label{shan_block}
\end{figure}

\begin{figure}[ht]
\centering
\begin{tabular}{cc}
{\includegraphics[width=5cm,height=5cm]{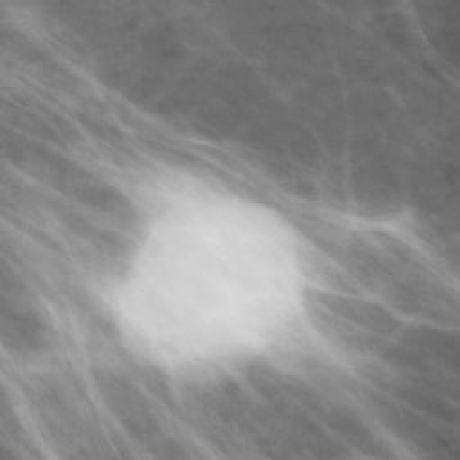}} \ & \
{\includegraphics[width=5cm,height=5cm]{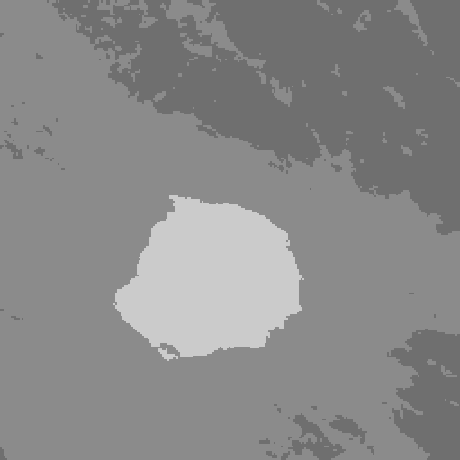}}\\
a) & b)
\end{tabular} 
\caption{The image mammography (\textit{a}). The segmented image with three gray levels (\textit{b}).}
\label{mammography}
\end{figure}

\begin{figure}
\centering
\includegraphics[width=6.66cm,height=5cm]{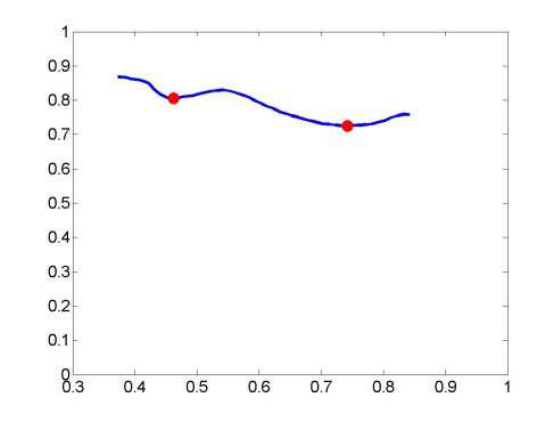}
\caption{The graphic of neutrosophic Shannon entropy where the
red circles represent the gray level thresholds (local minima) for image mammography.}
\label{shan_mammography}
\end{figure}

\begin{figure}[ht]
\centering
\begin{tabular}{cc}
{\includegraphics[width=5cm,height=5cm]{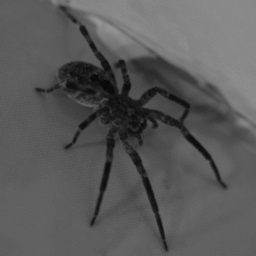}} \ & \
{\includegraphics[width=5cm,height=5cm]{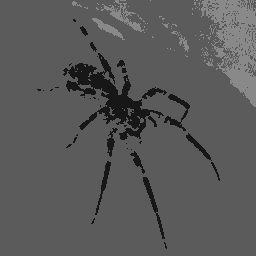}}\\
a) & b)
\end{tabular} 
\caption{The image spider (\textit{a}). The segmented image with three gray levels (\textit{b}).}
\label{spider}
\end{figure}

\begin{figure}[ht]
\centering
\includegraphics[width=6.66cm,height=5cm]{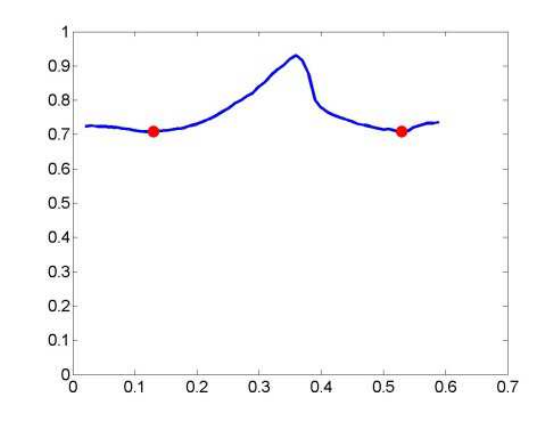}
\caption{The graphic of neutrosophic Shannon entropy where the red
circles represent the gray level thresholds (local minima) for image spider.}
\label{shan_spider}
\end{figure}

The proposed method was applied for segmentation of the following four
images: ball, block, mammography and spider. The obtained results can be
seen in figs. \ref{ball}, \ref{block}, \ref{mammography} and \ref{spider} 
while the entropy functions can be seen in figs. \ref{shan_ball},
 \ref{shan_block}, \ref{shan_mammography} and \ref{shan_spider}.
%%%%%%%%%%%%%%%%%%%%%
%%%%%%%%%%%%%%%%%%%%%

%%%%%%%%%%%%%%%%%%%%%%%%%%%%%%%%

%%%%%%%%%%%%%%%%%%%%%%%%%%%%%%%%%%%%%%%%%%%%%%%

%%%%%%%%%%%%%%%%%%%%%%%%%%%%%%%
 \section{Conclusions}

Based on the concept of neutrosophic information and the definition of
the degree of truth, degree of neutrality and degree of falsity, a new
image thresholding method is proposed. It utilizes the Shannon entropy
in order to determine adequate threshold values. It is expected that the
absolute value of the difference between degree of truth and degree of
falsity for each pixel can be as close to 1 as possible, so that the
entropy of each pixel is as minimal as possible. The proposed method
which is based on minimizing the neutrosophic entropy of an image has
demonstrated performance in multilevel thresholding. At the same time,
the experimental results indicate that the proposed method can find
appropriate threshold values.

%\newpage

\section*{References}

\begin{enumerate}

\item\label{r1} K. Atanassov. Intuitionistic Fuzzy sets. Fuzzy Sets and
Systems 20, pp. 87-96, (1986).
\item\label{r2} C. Ashbacher, Introduction to Neutrosophic Logic,
American Research Press, Rehoboth, NM, (2002).
\item\label{r3} V. Patrascu, Shannon entropy for imprecise and
under-defined or over-defined information, The 25th Conference on
Applied and Industrial Mathematics, CAIM 2017, Iasi, Romania, ROMAI
Journal, vol. 14, no. 1, pp. 169-185, (2018).
\item\label{r4} V. Patrascu, Shannon Entropy for Neutrosophic Information, doi:10.13140 \\
/RG.2.2.32352.74244, arXiv:1810.00748, September, (2018).
\item\label{r5}V. Patrascu, A Novel Penta-Valued Descriptor for Color
Clustering, The 6th International Conference on Image and Signal
Processing, ICISP 2014, Cherbourg, Normandy, France, June 30 - July 2,
2014, Volume: Image and Signal Processing, Lecture Notes in Computer
Science, Volume 8509, Springer International Publishing Switzerland, pp.
173-182, (2014).
\item\label{r6}V. Patrascu, New Framework of HSL System Based Color
Clustering Algorithm, The 24th Midwest Artificial Intelligence and
Cognitive Sciences Conference, MAICS 2013, April 13-14, 2013, New
Albany, Indiana. USA, ceur-ws.org, Vol. 1348, pp. 85-91, (2013).
\item\label{r7} U. Rivieccio, Neutrosophic Logics: Prospects and
Problems, Fuzzy Sets and Systems, 159, pp. 1860-1868, (2008).
\item\label{r8} F. Smarandache, A Unifying Field in Logics: Neutrosophic
Logic. Neutrosophy, Neutrosophic Set, Neutrosophic Probability, American
Research Press, Rehoboth, NM, (1999).
\item\label{r9} F. Smarandache, Neutrosophic Set - A Generalization of
the Intuitionistic Fuzzy Set, International Journal of Pure and Applied
Mathematics, 24, no. 3, pp. 287-297, (2005).
\item\label{r10} C.E. Shannon, A mathematical theory of communication,
Bell System Tech., J. 27, pp. 379-423, (1948).
\item\label{r11} Wikipedia, https://en.wikipedia.org/wiki/Multiset.
\item\label{r12} A. L. Zadeh. Fuzzy sets, Information and Control, 8,
pp. 338-353, (1965).

\end{enumerate}

\end{document}